\documentclass[a4paper]{article}

\usepackage[margin=1.25in]{geometry}

\usepackage{graphicx}
\usepackage{xcolor} 

\usepackage{amssymb}
\usepackage{amsmath}
\usepackage{amsthm}
\usepackage{bm}

\usepackage{booktabs}
\usepackage{multirow}
\usepackage{diagbox}

\usepackage{soul}
\usepackage{tcolorbox}
\usepackage{float}

\usepackage{hyperref}

\numberwithin{equation}{section}

\theoremstyle{definition}

\begin{document}

\title{YOLOv8-SMOT: An Efficient and Robust Framework for Real-Time Small Object Tracking via Slice-Assisted Training and Adaptive Association}

\author{
    Xiang Yu$^{1}$\thanks{Equal contribution.} \quad\quad
    Xinyao Liu$^{2}$\footnotemark[1] \quad\quad
    Guang Liang$^{1}$\footnotemark[1] \thanks{Corresponding author.} \\\\
    $^1$School of Artificial Intelligence, Nanjing University, China \\
    $^2$University of Science and Technology of China, Hefei, China \\\\
    {\tt\small 221300049@smail.nju.edu.cn, liuxinyao@mail.ustc.edu.cn, liangg@lamda.nju.edu.cn}
}

\maketitle

\begin{abstract}
\noindent
Tracking small, agile multi-objects (SMOT), such as birds, from an Unmanned Aerial Vehicle (UAV) perspective is a highly challenging computer vision task. The difficulty stems from three main sources: the extreme scarcity of target appearance features, the complex motion entanglement caused by the combined dynamics of the camera and the targets themselves, and the frequent occlusions and identity ambiguity arising from dense flocking behavior. This paper details our championship-winning solution in the MVA 2025 "Finding Birds" Small Multi-Object Tracking Challenge (SMOT4SB), which adopts the tracking-by-detection paradigm with targeted innovations at both the detection and association levels. On the detection side, we propose a systematic training enhancement framework named \textbf{SliceTrain}. This framework, through the synergy of 'deterministic full-coverage slicing' and 'slice-level stochastic augmentation, effectively addresses the problem of insufficient learning for small objects in high-resolution image training. On the tracking side, we designed a robust tracker that is completely independent of appearance information. By integrating a \textbf{motion direction maintenance (EMA)} mechanism and an \textbf{adaptive similarity metric} combining \textbf{bounding box expansion and distance penalty} into the OC-SORT framework, our tracker can stably handle irregular motion and maintain target identities. Our method achieves state-of-the-art performance on the SMOT4SB public test set, reaching an SO-HOTA score of \textbf{55.205}, which fully validates the effectiveness and advancement of our framework in solving complex real-world SMOT problems. The source code will be made available at \url{https://github.com/Salvatore-Love/YOLOv8-SMOT}.
\end{abstract}

\section{Introduction}

With the proliferation of Unmanned Aerial Vehicle (UAV) technology in the field of autonomous systems, its application as an aerial perception platform has become increasingly widespread, particularly in areas such as ecological monitoring, agricultural inspection, and public safety \cite{kondo2023mva2023, liu2021survey}. While UAVs offer unprecedented flexibility and perspectives for capturing wide-area scenes, they also pose severe challenges to computer vision algorithms. Among these applications, the continuous localization and identity maintenance of multiple small, moving targets in video sequences, known as Small Multi-Object Tracking (SMOT), is a particularly critical and arduous fundamental task.

The SMOT task from a UAV perspective, especially when tracking agile creatures like birds, is far more complex than traditional Multi-Object Tracking (MOT) scenarios \cite{milan2016mot16, zhu2018vision}. This complexity arises from the interplay of three core challenges:
\begin{itemize}
    \item \textbf{Extreme Scarcity of Appearance Information:} Targets often occupy only a few dozen pixels, containing almost no distinguishable texture or color features. This fundamentally invalidates classic tracking paradigms that rely on appearance models for re-identification (Re-ID), such as DeepSORT \cite{wojke2017simple}.
    \item \textbf{Complex Motion Entanglement \cite{yin2020unified}:} The tracker must not only handle the target's free, non-linear movement in three-dimensional space but also contend with the drastic camera motion caused by the UAV's own complex translations, rotations, and altitude changes. The superposition of these two motion patterns results in extremely complex and unpredictable apparent motion of the target in the image plane, causing frequent tracking failures for traditional methods that rely on linear motion models like the Kalman filter \cite{bewley2016simple}.
    \item \textbf{Dense Flocking Dynamics \cite{reynolds1987flocks}:} The unique flocking behavior of birds leads to frequent and severe occlusions among targets, with high similarity in appearance and motion patterns between individuals. This poses an extreme test for maintaining individual identity continuity solely through motion information without relying on appearance features, making it highly susceptible to numerous Identity Switch (ID Switch) errors.
\end{itemize}

Although significant progress has been made in general MOT tasks with paradigms based on tracking-by-detection \cite{bewley2016simple,wojke2017simple}, joint detection and tracking \cite{wang2020towards,zhang2021fairmot}, and Transformers \cite{sun2020transtrack,zeng2022motr}, they struggle to directly address the cumulative effect of the three challenges mentioned above. To systematically advance this field, MVA 2025 organized the "Finding Birds" Small Multi-Object Tracking Challenge (SMOT4SB) \cite{baselinecode_mva2025_smot4sb_challenge}, providing the first large-scale dataset designed for such extreme scenarios and a new evaluation metric, SO-HOTA \cite{mva2025_smot4sb_challenge}.

This paper presents our championship-winning solution for this challenge. We believe that to overcome this problem, it is essential to build a framework that can synergistically address the bottlenecks of both detection and association. To this end, we propose an efficient tracking-by-detection system whose core contributions lie in two aspects, each precisely addressing the aforementioned challenges:

\begin{enumerate}
    \item \textbf{Detector Optimization for Information Scarcity:} To fundamentally solve the problem of detecting small objects, we propose a systematic training data enhancement framework called \textbf{SliceTrain}. Through a two-stage process combining "deterministic full-coverage slicing" and "slice-level stochastic augmentation," this framework significantly enriches the diversity and information density of training samples without sacrificing information integrity. This allows the detector (YOLOv8) to be trained efficiently with a larger batch size under limited computational resources, thereby significantly enhancing its feature capture and localization capabilities for tiny objects \cite{ozge2019power}.
    \item \textbf{Robust Tracker for Complex Dynamics:} To address the association challenges posed by motion entanglement and flocking behavior, we designed a tracker that is completely independent of appearance features. We have deeply enhanced the OC-SORT \cite{cao_observation-centric_2023} framework by introducing a \textbf{motion direction maintenance} mechanism to smooth out noise from non-linear motion and designing an \textbf{adaptive similarity metric}. This metric combines bounding box expansion and distance penalty, effectively solving the matching difficulties caused by the small size of targets and frequent close-range intersections.
\end{enumerate}

Our method achieves state-of-the-art performance on the SMOT4SB dataset, reaching an SO-HOTA score of \textbf{55.205} on the public test set, which validates the advancement and effectiveness of our framework in solving complex real-world SMOT tasks.

\section{Related Work}

\noindent\textbf{Multi-Object Tracking (MOT).} The predominant paradigm in the MOT field is "tracking-by-detection" \cite{bewley2016simple,wojke2017simple}, which decouples detection and association, allowing for independent optimization of each module. Classic algorithms like SORT \cite{bewley2016simple} and DeepSORT \cite{wojke2017simple} use a Kalman filter for motion prediction and combine it with IoU or appearance features for data association. In recent years, researchers have proposed joint detection and tracking frameworks \cite{wang2020towards,zhang2021fairmot} and end-to-end Transformer architectures \cite{sun2020transtrack} to improve efficiency and accuracy. However, these general MOT methods mostly rely on datasets like the MOTChallenge series \cite{milan2016mot16}, where the primary targets are pedestrians or vehicles, which are typically large and have distinct appearance features, a significant difference from SMOT scenarios.

\noindent\textbf{Small Object Detection (SOD).} Small Object Detection (SOD) aims to solve the problem of limited appearance cues due to the small size of the target \cite{liu2021survey}. Traditional methods improve performance through multi-scale feature fusion (such as Feature Pyramid Networks, FPN \cite{lin2017feature}) and data augmentation \cite{zoph2020learning}. The demands on detectors are particularly stringent in scenes like aerial imagery, where targets are tiny and dense. The emergence of benchmark datasets like VisDrone \cite{zhu2018vision} has spurred development in this area. At the data level, while traditional Random Cropping can increase data diversity, its sampling process is stochastic. For sparse small objects in high-resolution images, it may fail to sample them effectively over many iterations, leading to insufficient information utilization. Methods like Slicing Aided Hyper Inference (SAHI) \cite{akyon2022slicing} mainly focus on using slicing to assist inference, with their training-stage slicing being relatively straightforward.

In contrast, our proposed SliceTrain framework focuses on constructing a superior training paradigm. It is not a simple random crop but ensures 100\% utilization of the original data through systematic full-coverage slicing. It is not merely for data partitioning but applies fine-grained augmentation transformations after slicing to create a training set with high information density and diversity. This design aims to resolve the inherent conflict between comprehensiveness and diversity in data utilization found in traditional methods.

\noindent\textbf{Small Multi-Object Tracking (SMOT).} The core challenge of the SMOT task is how to perform reliable cross-frame association when appearance features are virtually unusable \cite{liu2021survey}. Existing SMOT datasets, such as UAVDT \cite{du2018unmanned} and VisDrone \cite{zhu2018vision}, mostly focus on targets with constrained motion in urban environments. However, the SMOT4SB dataset is the first to systematically introduce the challenge of "motion entanglement" \cite{yin2020unified}, where both the camera and the target move freely in three-dimensional space. This complex relative motion pattern makes it extremely difficult to rely solely on motion prediction, placing higher demands on the robustness of the tracker. This prompted us to design a tracking algorithm that does not rely on appearance information but instead deeply mines and utilizes motion consistency and similarity metrics.

\section{Method}
Our tracking framework is designed according to the tracking-by-detection paradigm, which decouples detecting and tracking matching procedures. This paradigm allows us to optimize the detector and tracker separately to build an almost optimal MOT model.

\begin{figure}[htb]
    \centering
    \includegraphics[width=0.98\linewidth]{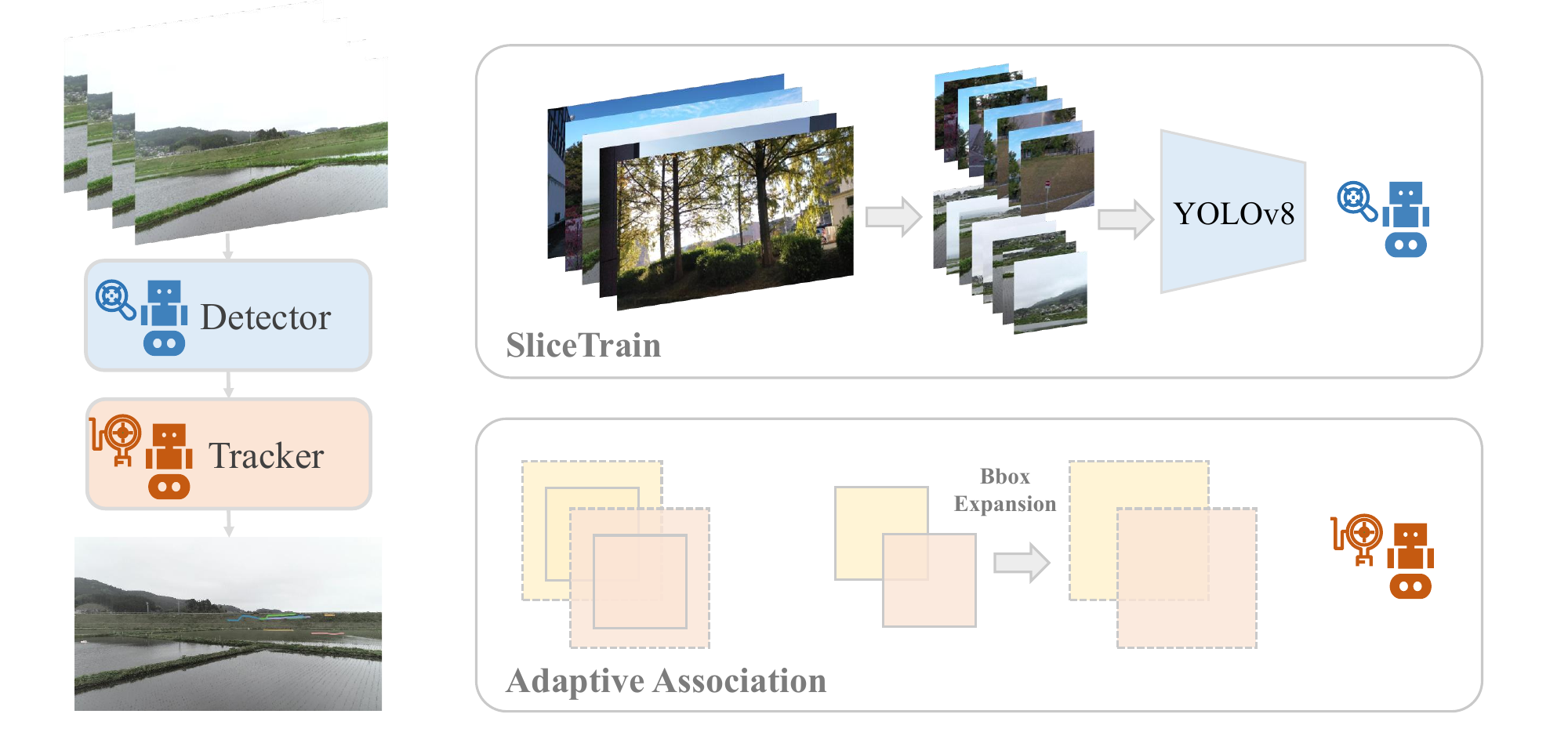}
    \caption{Model Overview.}
    \label{fig:overview}
\end{figure}

\subsection{Detector}
Our detection module is based on the powerful YOLOv8 model, and its leap in performance is primarily attributed to our designed SliceTrain framework. This framework preprocesses the data before training, with its core mechanism divided into two key steps: first, enhancing the model's perception of minute details through high-quality fine-tuning, and then performing efficient inference on the original full-size images.

\subsubsection{SliceTrain Framework: Overcoming the Resolution-Diversity Dilemma}
High-quality small object detection, when dealing with high-resolution images, generally faces a core predicament: \textbf{The Resolution-Diversity Dilemma}. On one hand, high-resolution input (e.g., $3840 \times 2160$) is key to capturing the details of tiny objects. On the other hand, the enormous memory overhead limits the batch size to extremely small values (e.g., 1 or 2), leading to monotonous training samples, unstable gradient updates, and difficulty for the model to learn generalizable features \cite{ozge2019power, akyon2022slicing}.

To overcome this dilemma, we designed \textbf{SliceTrain}, a systematic training data enhancement framework. It is not a simple slicing operation but constructs a data stream with extremely high information density and diversity through two complementary core steps: \textbf{Deterministic Full-Coverage Tiling} and \textbf{Slice-Level Stochastic Augmentation}.

\noindent\textbf{Step 1: Deterministic Full-Coverage Tiling}
This principle aims to losslessly decompose high-resolution images into model-processable units while ensuring information integrity. We use overlapping sliding windows to deterministically segment each high-resolution image into a set of "tiles" (e.g., sliced into $1280 \times 1280$ tiles with a certain overlap ratio). Unlike random cropping, this \textbf{deterministic} grid division ensures that \textbf{every pixel} of the original image is covered by at least one tile, achieving \textbf{lossless utilization} of spatial information. The \textbf{overlapping design} ensures that objects located at the slice boundaries are fully contained in at least one tile, fundamentally preventing the loss of target information due to segmentation.

\noindent\textbf{Step 2: Slice-Level Stochastic Augmentation}
This principle aims to inject maximum data diversity into the model. After obtaining the full-coverage tile set, the framework treats each tile as an independent image sample and applies a series of strong random data augmentations (e.g., Mosaic, color jitter, random geometric transformations). The key is that the augmentation is applied independently at the \textbf{slice level}. This means that when constructing a training batch, the model sees not only tiles from different source images but also samples that have undergone different visual transformations.

\noindent\textbf{Framework Efficacy: Constructing High-Density, High-Diversity Training Batches}
The final output of the SliceTrain framework is a high-quality training data stream that surpasses traditional methods in both \textbf{information density} and \textbf{scene diversity}. When the model samples a batch from this stream, it is no longer exposed to a few complete, relatively sparse images, but rather a \textbf{high-density sample set} composed of slices from different source images, different spatial locations, and subjected to different visual transformations. This design transforms each gradient update into an information-rich, multi-dimensional learning event, fundamentally accelerating model convergence and enhancing its generalization ability to complex real-world scenarios.

\subsubsection{Full Size Inference}
Although the model is trained on the SliceTrain framework, during the inference stage, we apply it directly to the original, unsliced full-size test images.

This asymmetric strategy of "slice for training, full-size for inference" is key to our detector's efficiency and high accuracy. It cleverly bypasses the complex process of slicing, predicting, and then stitching at inference time, thus avoiding additional computational overhead and potential errors introduced by image stitching. By being fine-tuned with high intensity on sub-images, our model becomes exceptionally sensitive to small objects. When it performs inference on a full-size image, this "magnified" perceptual ability allows it to accurately locate those tiny targets that are easily overlooked in a vast background. This method preserves the global context of the scene while fully leveraging the advantages of fine-grained training, ultimately forming a detection process that is both fast and accurate.

\subsection{Tracker}
\subsubsection{Preliminaries}
Our proposed tracker is an enhancement of the Observation-Centric SORT (OC-SORT) framework \cite{cao_observation-centric_2023} integrated with ByteTrack \cite{zhang_bytetrack_2022}, which achieve matching without appearance features as they are often unreliable when the target object is too small. To contextualize our contributions, we first briefly outline the preliminary concepts of OC-SORT, which improves upon traditional trackers by employing a multi-stage strategy that prioritizes detector observations over model predictions, especially during challenging scenarios.

To simplify the illustration, we use $T$ to denote the possible tracking detections and divide them into 3 types according to detection confidence score:
\[
\begin{aligned}
\tau_\text{high} &= \{t\mid t\in T \text{ and score}(x) \ge \text{threshold}_\text{track-high}\}\\
\tau_\text{low} &= \{t\mid t\in T \text{ and threshold}_\text{high} > \text{score}(x) \ge \text{threshold}_\text{low}\}\\
\tau_\text{discard} &= \{t\mid t\in T \text{ and score}(x) < \text{threshold}_\text{low}\}
\end{aligned}
\]

OC-SORT refines the standard tracking-by-detection paradigm through a sequence of specialized filtering and updating stages:
\begin{itemize}
    \item Enhanced Association: In the primary matching step, OC-SORT supplements the standard IoU cost with an Observation-Centric Momentum (OCM) for matching $\tau_\text{high}$. This cost is derived from the motion direction calculated using historical observations, providing a more stable association cue than the noisy velocity estimate from the Kalman Filter.
    \item Secondary Matching: In the next matching step, a tracker focuses on $\tau_\text{low}$. These are typically tracks whose objects have become occluded. The association is performed using only IoU as the similarity metric. This step is crucial for maintaining trajectory continuity, as it effectively links tracks through periods of partial or full occlusion by "rescuing" the low-score but valid detections. Detections except $\tau_\text{high}$ that remain unmatched after this stage are discarded as likely background noise.
    \item Heuristic Recovery: As a final step, an Observation-Centric Recovery (OCR) stage performs a second, simpler matching attempt for remaining unmatched tracks with relatively high confidence. It tries to associate any remaining unmatched tracks and detections based on the tracks' last known observations, effectively recovering objects that may have stopped or were briefly occluded.
\end{itemize}

\begin{figure}[htb]
    \centering
    \includegraphics[width=0.9\linewidth]{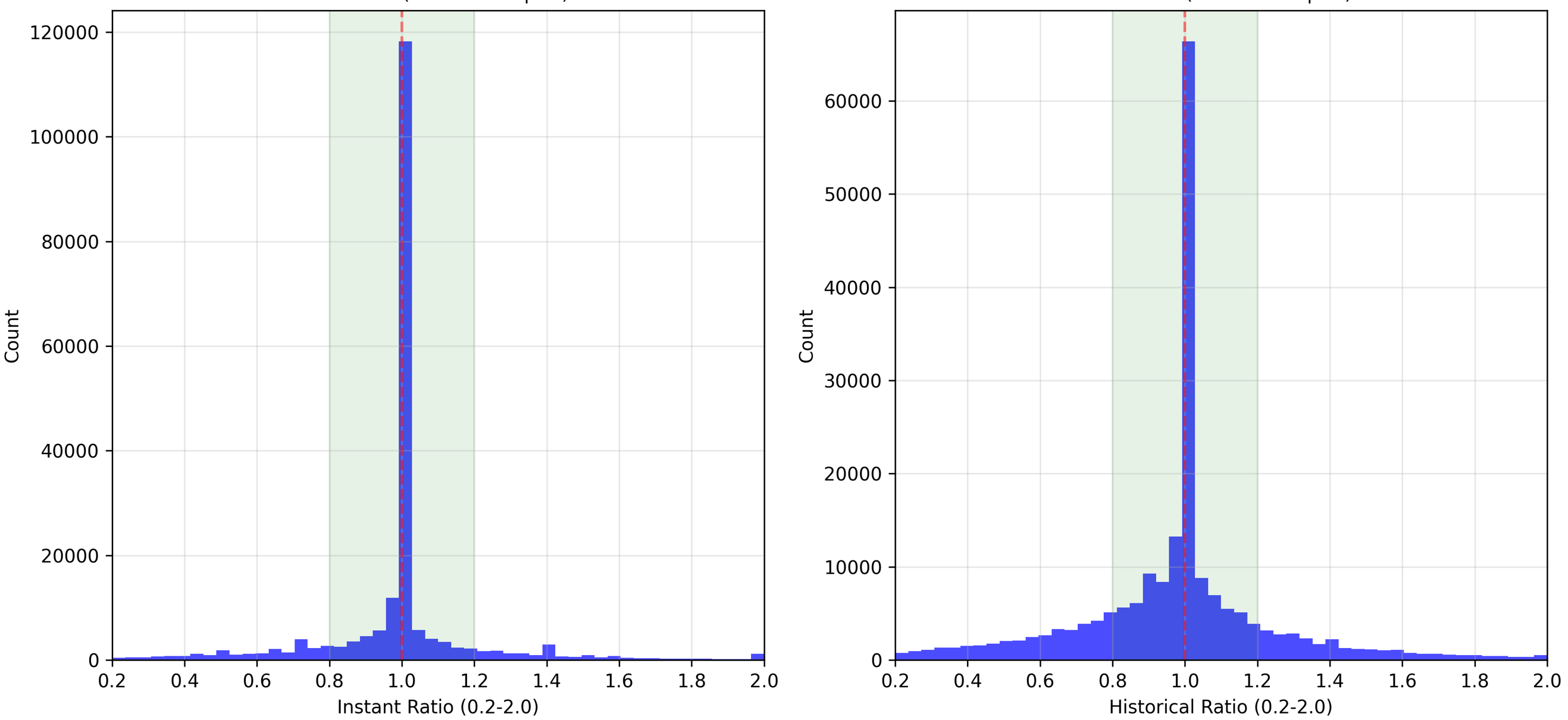}
    \caption{Distribution of velocity of all instances of training dataset. The instance ratio is $\frac{\text{speed between frame }t\text{ and }t-1}{\text{speed between frame }t-1\text{ and }t-2}$. The historical ratio is $\frac{\text{speed between frame }t\text{ and frame }t-1}{\text{average speed over last }4\text{ frame}}$}.
    \label{fig:velocity analysis}
\end{figure}

Though OC-SORT assume that the tracking targets have a constant velocity within a time interval, which is called the linear motion assumption, we discover almost the velocity of $76.2\%$ annotated bird instances change no more than $\pm 20\%$ compared with the previous frame and $64.2\%$ of the previous $4$ frames as Figure \ref{fig:velocity analysis} shows. More results can be found in Table \ref{tab:velocity analysis}. Therefore we can regard most of bird movements satisfy the linear motion assumption and OC-SORT is a reasonable choice for this MOT task.

\begin{table}[htb]
    \centering
    \setlength{\tabcolsep}{14pt}
    \begin{tabular}{cc}
        \toprule
        \textbf{Reference Window (N Frames)} & \textbf{Samples with Ratio in [0.8, 1.2]} \\
        \midrule
        $1$ & $76.2\%$ \\
        $2$ & $71.9\%$ \\
        $3$ & $67.3\%$ \\
        $4$ & $64.2\%$ \\
        $5$ & $61.2\%$ \\
        \bottomrule
    \end{tabular}
    \caption{Analysis of the constant velocity assumption on the training dataset. The table shows the percentage of samples where the velocity ratio $\frac{V}{\overline{V}}$ falls within the range $[0.8, 1.2]$, indicating near-uniform motion. The reference velocity $\overline{V}$ is calculated using a window of $N$ preceding frames.}
    \label{tab:velocity analysis}
\end{table}

\subsubsection{Motion Direction Maintaining}
The lack of information of appearance features, the small object tracking is a challenge task. To alleviate this problem, we need to take fully advantage of motion features. However, the characteristic of bird locomotion is the absence of regularity, reflected both in the speed and direction of movement. This implies model the motion may require quite a lot of parameters, as it's also hard for humans to predict the next moment's movement of a bird in the distance. This will leads to significant real-time performance degradation.

As a trade-off, we propose to apply Exponential Moving Average (EMA) technique to maintain historical velocity direction. We argue that the historical velocity maintained by EMA is a better choice than simply using the kth frame before target frame to calculate a more precise cosine direction cost. The extra costs EMA incurs will not be a burden for real-time application and it can avoid the hacking of a sudden turn. By denoting the instant velocity as $v_{\text{ins}}^t$ and the historical EMA velocity as $v_{\text{EMA}}^t$ at frame t, the updating procedure is formulated as:
\begin{equation}
    v_{\text{EMA}}^t = \alpha v_{\text{EMA}}^{t-1} + (1 - \alpha) v_{\text{ins}}^{t-1}
    \label{eq:EMA}
\end{equation}

\subsubsection{Similarity Metric Adapting}
\begin{figure}[htb]
    \centering
    \includegraphics[width=0.3\linewidth]{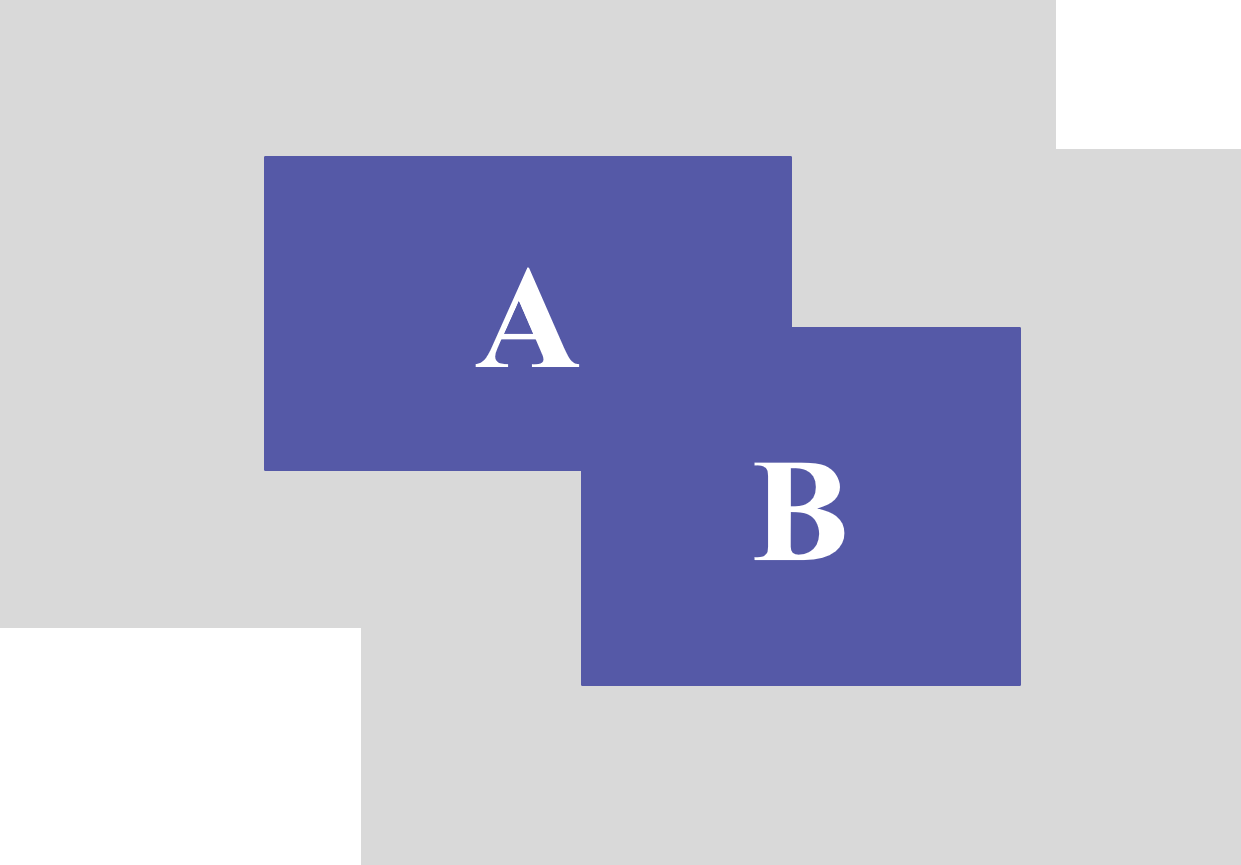}
    \caption{Example of IoU caculation of object A and object B after expanding bounding boxes. The blue part is the detected or predicted bounding box, and the gray part is expanded bounding box. The IoU calculation is based on the expanded bounding boxes, i.e., gray bounding boxes.}
    \label{fig:bbox expansion}
\end{figure}

The default similarity metric of OC-SORT is IoU, which becomes invalid when two bounding boxes are disjoint and this situation often occurs on small objects. This property of small objects makes matching difficult during tracking. But we propose to utilize this characteristic: considering that object is small, the possibility of overlapping and large shifts is relatively lower than certain MOT tasks. Based on this assumption, we expand the size of bounding box to simulate an ordinary objects matching task before calculating IoU. The main idea of bounding box expansion is illustrated in Figure \ref{fig:bbox expansion} and it's similar to \cite{yang_hard_2023}.

Additionally, we extend the IoU metric with distance as a penalty term in consideration of that the object does not exhibit significant displacement between frames, which is similar to DIoU \cite{zheng_distance-iou_2019} actually. To normalize this metric, it can be expressed as follows:
\begin{equation}
\text{Similarity} = \frac{\text{ExpandedIoU} - \text{NormalizedDistance} + 1}{2}
\end{equation}

\section{Experiments}
In this section, we evaluate our tracker with YOLOv8 of different parameters. We compare effects of several detectors, plus ablation studies to verify the robustness and generalization ability of our improvements in different scenarios.

\subsection{Datasets and Metrics}
\textbf{Datasets.} We conducted all experiments on SMOT4SB \cite{mva2025_smot4sb_challenge} dataset in which most objects are small and process irregular motion patterns. SMOT4SB is a dataset comprised of videos captured by UAVs, consisting of 128 training sequences, 38 validation sequences, and 45 testing sequences. Different from ordinary MOT dataset \cite{milan_mot16_2016,dendorfer_mot20_2020}, SMOT4SB possesses higher difficulties including: 1) targets’ irregular motions, 2) sudden and large camera movements, and 3) limited appearance information of small objects.

\textbf{Metrics.} In experiments, we select the official metric SO-HOTA metrics (i.e., SO-HOTA, SO-DetA and SO-AssA) \cite{mva2025_smot4sb_challenge} adopted from HOTA metrics (i.e., HOTA, DetA and AssA) \cite{luiten_hota_2021}. SO-HOTA introduces Dot Distance (DotD) \cite{xu_dot_2021} for similarity scoring, which compares precise point-like object representations. 

\subsection{Implementation Details}
\textbf{Detector.} Our detector YOLOv8-SOD is built upon three different sizes (L, M, S) of YOLOv8 \cite{Jocher_Ultralytics_YOLO_2023}. We employ the SliceTrain strategy for fine-tuning: original high-resolution training images (e.g., $2160\times 3840$) are sliced into $1280\times 1280$ overlapping sub-images with an overlap ratio of 20\%. This strategy allows the training batch size on a single Nvidia RTX 3090 GPU to be effectively increased from 1 (for full images) to 6 (for sliced images). All models are trained on the training set, and inference is performed directly on the original full-size images to ensure efficiency and global context.

\begin{figure}[htb]
    \centering
    \includegraphics[width=0.9\linewidth]{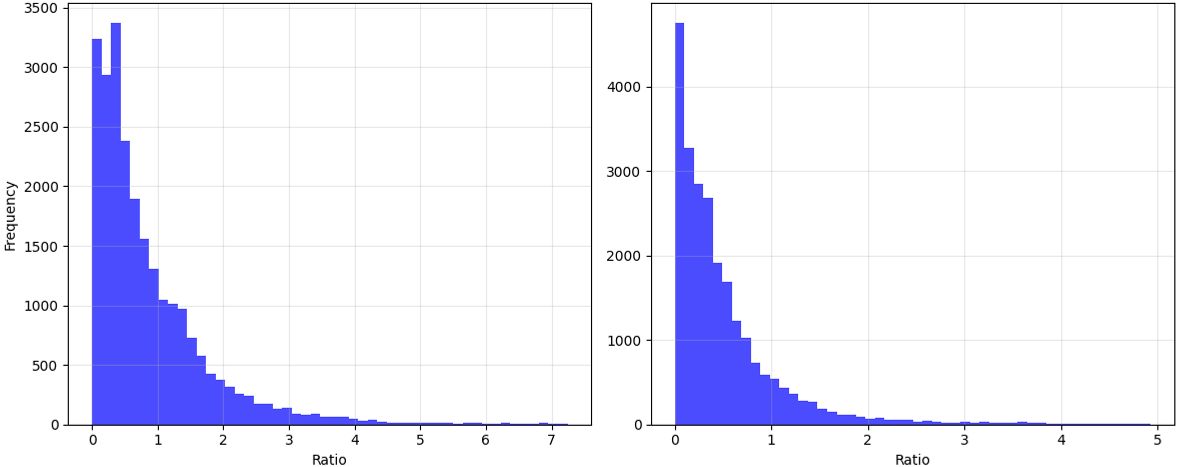}
    \caption{The ratio of horizontal (left) or vertical (right) displacement to width (left) or height (right) of bounding boxes in the training dataset.}
    \label{fig:bbox analysis}
\end{figure}

\textbf{Tracker.} To determine the optimal values for the IoU threshold used in matching and the hyperparameter associated with the reduction of this threshold for stages 2 and 3 of matching, we employed a grid search approach. This process identified an optimal IoU threshold of $0.25$ and a decrement of $0.08$. Regarding the track threshold concerning confidence score, we established it at $0.25$, based on the observation from our detection results on training dataset. We set $\alpha = 0.8$ of EMA of \ref{eq:EMA} by cross validation on training dataset. An expansion scale of 2 was chosen for simplicity. This decision stems from our empirical observation: when displacements less than $30\%$ of the bounding box's shorter dimension are disregarded as relatively still movements, the inter-frame x- and y-axis displacements of the majority of targets' bounding boxes do not exceed twice their respective width and height. Figure \ref{fig:bbox analysis} provides a statistical illustration of these bounding box movements.

\subsection{Benchmark Evaluation}
We conduct the experiments on the public testing dataset. We compare different detector with the same tracker in table \ref{tab:results of different detectors} to demonstrate that our tracker can achieve satisfactory performance and the capacity of real-time task. 

\begin{table}[h!]
    \centering
    \setlength{\tabcolsep}{9.7pt}
    \begin{tabular}{cccccc}
        \toprule 
        Model & SO-HOTA & SO-DetA & SO-AssA & Memory (MiB) & Speed (FPS) \\
        \midrule 
        YOLOv8-L & \textbf{55.205} & \textbf{51.716} & \textbf{59.082} & 5036 & 5.70 \\
        YOLOv8-M & 54.426 & 49.529 & 59.962 & 3884 & 8.96 \\
        YOLOv8-S & 53.808 & 48.388 & 59.979 & 3190 & \textbf{17.61} \\
        Baseline & 10.676 & 9.788 & 11.671 & \textbf{2346} & 8.88\\
        \bottomrule
    \end{tabular}
    \caption{Comparison of different detectors on the public testing dataset of SMOT4SB. The memory occupied and inference speed are measured on single Nvidia RTX 3090 GPU with resolution of $2160\times 3840$. The baseline model \cite{baselinecode_mva2025_smot4sb_challenge} is tested with resolution of $1792\times 3264$. The best results are shown in bold.}
    \label{tab:results of different detectors}
\end{table}

As presented in Table \ref{tab:results of different detectors}, our evaluation demonstrates a clear trade-off between model performance and computational efficiency across different scales of the YOLOv8 detector. The largest model, YOLOv8-L, achieves the highest tracking accuracy (SO-HOTA of $55.205$) but at the lowest speed ($5.70$ FPS). In contrast, the smallest model, YOLOv8-S, boosts the inference speed nearly threefold to $17.61$ FPS and reduces memory usage significantly. This substantial efficiency gain is achieved with only a marginal performance drop of $1.397$ in SO-HOTA, validating its suitability for applications requiring near real-time performance.

Furthermore, the efficiency of our model can be significantly enhanced through model quantization. By applying advanced Quantization-Aware Training (QAT) methods like GPLQ \cite{liang2025gplq}, or practical Post-Training Quantization (PTQ) techniques such as QwT \cite{fu2025quantization} and QwT-v2 \cite{tang2025qwt}, the computational footprint can be further reduced. This would enable our high-performance tracking framework to be deployed on low-power edge devices while maintaining real-time processing capabilities.

\subsection{Ablations}
To validate the effectiveness of our proposed enhancements, we conducted a comprehensive ablation study. A primary challenge in tracking small objects is the unreliability of the standard Intersection over Union (IoU) metric, which often fails even when objects have only minor displacements between frames.

\begin{table}[h!]
    \centering
    \setlength{\tabcolsep}{15pt}
    \begin{tabular}{ccccc}
        \toprule
        Percentile & Default & + BBox Expansion & + Center Distance & Both \\
        \midrule
        10 & 0.048 & 0.297 & 0.445 & \textbf{0.621} \\
        30 & 0.326 & 0.550 & 0.634 & \textbf{0.766} \\
        50 & 0.531 & 0.702 & 0.754 & \textbf{0.848} \\
        70 & 0.745 & 0.836 & 0.870 & \textbf{0.917} \\
        90 & 0.996 & 0.997 & 0.998 & \textbf{0.999} \\
        \bottomrule
    \end{tabular}
    \caption{Statistical Results for Different IoU Calculation Methods on the Training Set. The bounding boxes where the displacement of adjacent frame is no more than $30\%$ of the shorter dimension are not included.}
    \label{tab:iou methods analysis}
\end{table}

To precisely quantify this issue, we performed a statistical analysis focused on challenging but common tracking scenarios. We filtered the training set for pairs of ground-truth bounding boxes where the inter-frame displacement was less than $30\%$ of the shorter box dimension, representing objects that are relatively stable yet difficult for standard IoU. The results, shown in Table \ref{tab:iou methods analysis}, are striking. The Default IoU metric struggles immensely, yielding a score of only 0.048 at the 10th percentile. In stark contrast, when using both our BBox Expansion and Center Distance methods, the 10th percentile score rises to 0.621, and the median score reaches 0.848. This proves that our proposed metric provides a dramatically more robust and consistent similarity signal under the exact conditions where standard IoU fails.

\begin{table}[h!]
    \centering
    \setlength{\tabcolsep}{25.9pt}
    \begin{tabular}{cccc}
        \toprule
        Additional Method & SO-HOTA & SO-DetA & SO-AssA \\
        \midrule
        Default & 44.200 & 46.094 & 42.533 \\
        + EMA & 47.916 & 48.015 & 48.007 \\
        + BBox Expansion & 51.387 & \textbf{51.432} & 51.488 \\
        + Dist Penalty & \textbf{55.205} & 51.761 & \textbf{59.082} \\
        \bottomrule
    \end{tabular}
    \caption{The evolution trajectory of tracker. All models are evaluated on public dataset of SMOT4SB together with YOLOv8-L. The default tracker is OC-SORT. The best results are shown in bold.}
    \label{tab:evolution trajectory}
\end{table}

The direct impact of this robustified metric on overall tracking performance is detailed in Table \ref{tab:evolution trajectory}. Starting from a baseline SO-HOTA of 44.200, we incrementally added our contributions. After integrating Exponential Moving Average (EMA) for motion stability (raising SO-HOTA to 47.916), we introduced our full enhanced similarity metric, broken down into two steps. First, adding Bounding Box Expansion boosted SO-HOTA to 51.387. Second, adding the Distance Penalty provided the most substantial gain, elevating the final SO-HOTA to 55.205. This confirms that the robust similarity signal, specifically designed for and validated under challenging low-displacement conditions (as shown in Table \ref{tab:iou methods analysis}), directly translates into superior tracking accuracy.

\section{Conclusion}

In this paper, we presented YOLOv8-SMOT, our winning solution to the MVA 2025 SMOT4SB challenge for tracking small, agile objects from UAVs. Our tracking-by-detection approach features two key innovations: the \textbf{SliceTrain} framework for superior small object detection, and a robust, appearance-free tracker enhanced with \textbf{motion direction maintenance} and an \textbf{adaptive similarity metric}. Achieving a state-of-the-art SO-HOTA score of \textbf{55.205}, our method demonstrates strong performance. However, we recognize its reliance on motion heuristics as a limitation, particularly in cases of extreme motion or prolonged occlusion. Future work could thus explore more advanced dynamic models or techniques for minimal feature extraction. We hope this work provides a robust baseline and valuable insights for advancing research in the challenging SMOT field.

\bibliographystyle{plain}
\bibliography{references}
\end{document}